\documentclass[10pt, letterpaper, conference]{IEEEtran}
\IEEEoverridecommandlockouts
\usepackage{color}

\usepackage{times}
\usepackage{epsfig}
\usepackage{graphicx}
\usepackage{tabularx}
\usepackage{amsmath}
\usepackage{amssymb}
\usepackage{bbm}
\usepackage[misc]{ifsym}
\usepackage[export]{adjustbox}
\usepackage{amsthm}
\usepackage{float}

\usepackage[ruled,vlined,linesnumbered]{algorithm2e}
\usepackage{graphicx} %% This one is the package for using \includegraphics
\usepackage{subfigure}
\usepackage{tikz}
\usepackage{verbatim}
\usepackage{booktabs}
\usepackage{multirow}
\usepackage{makecell}
\usepackage{authblk}
\usepackage{hyphenat} % for preventing hyphenation, using environment \nohyphens{(text block)}

\usepackage{pgf}
\usepackage{algpseudocode}
\usetikzlibrary{arrows,automata}

\usepackage{graphbox}

\makeatletter
\newcommand{\removelatexerror}{\let\@latex@error\@gobble}
\makeatother

\makeatletter 
  \newcommand\figcaption{\def\@captype{figure}\caption} 
  \newcommand\tabcaption{\def\@captype{table}\caption} 
\makeatother

\makeatletter
\let\NAT@parse\undefined
\makeatother
\usepackage[colorlinks]{hyperref}  %hyperref still needs to be put at the end!

\usepackage{graphics}
\usepackage{pifont}
\usepackage{times}

\usepackage{soul}
\usepackage{url}
\usepackage{graphicx}
\usepackage{amsmath}
\usepackage{booktabs}
\usepackage[ruled]{algorithm2e}
\usepackage{balance} % for balancing columns on the final page
\usepackage[T1]{fontenc}    % use 8-bit T1 fonts
\usepackage{hyperref}
\usepackage{wrapfig}
\usepackage{subfigure}
\usepackage{fancybox}
\usepackage{tikz}
\usepackage{caption}
\usepackage{algpseudocode}
\usepackage{algorithmicx}
\usepackage[ruled,vlined,linesnumbered]{algorithm2e}
\usepackage{diagbox}
\usepackage{multirow}
\usepackage{amsfonts}       % blackboard math symbols
\usepackage{nicefrac}       % compact symbols for 1/2, etc.
\usepackage{microtype}      % microtypography

\usepackage{graphbox}

\usepackage{hyperref}

\graphicspath{{figures/}}

\begin{document}

\title{\Large \textbf{Game-theoretic Utility Tree for Multi-Robot Cooperative Pursuit Strategy}}

\author{Qin Yang \and Ramviyas Parasuraman
\thanks{The authors are from the Heterogeneous Robotics Lab, Department of Computer Science, University of Georgia, Athens, GA 30602, USA. 

Email: {\it \{qy03103,ramviyas\}@uga.edu}. }
}

\maketitle

\begin{abstract}
Underlying relationships among multiagent systems (MAS) in hazardous scenarios can be represented as game-theoretic models. In adversarial environments, the adversaries can be intentional or unintentional based on their needs and motivations. Agents will adopt suitable decision-making strategies to maximize their current needs and minimize their expected costs. This paper proposes and extends the new hierarchical network-based model, termed Game-theoretic Utility Tree (GUT), to arrive at a cooperative pursuit strategy to catch an evader in the Pursuit-Evasion game domain. We verify and demonstrate the performance of the proposed method using the Robotarium platform compared to the conventional constant bearing (CB) and pure pursuit (PP) strategies. The experiments demonstrated the effectiveness of the GUT, and the performances validated that the GUT could effectively organize cooperation strategies, helping the group with fewer advantages achieve higher performance.
\end{abstract}


\section{Introduction}
\label{sec:intro}

Although cooperative multiagent system (MAS)\footnote{In this paper, we use the terms agent and robot interchangeably.} decision-making is studied by many research communities, such as evolutionary computation, complex systems, game theory, graph theory, and control theory, these problems are either be episodic or sequential \cite{russell2010artificial}. Agents' actions or behavior are usually generated from a sequence of actions or policies, and the decision-making algorithms are evaluated based on policy optimality, search completeness, time complexity, and space complexity.\footnote{A policy is optimal if it has the highest utility/reward. A search algorithm is complete if it guarantees to return an optimal policy in a finite time when it exists. Time complexity quantifies the amount of time needed to search for a solution, while space complexity quantifies the amount of required computational memory \cite{rizk2018decision}.}

From a game-theoretic control perspective, most of the research in game theory has been focused on single-stage games with fixed, known agent utilities \cite{browning2005stp}, such as distributed control in communication \cite{zhang2014game} and task allocation \cite{bakolas2021decentralized}. Especially, recent MAS research domains focus on solving path planning problems for avoiding static or dynamical obstacles \cite{agmon2011multi} and formation control \cite{shapira2015path} from the unintentional adversary perspective. For intentional adversaries, the "pursuit domain" \cite{chung2011search,kothari2014cooperative} primarily deals with how to guide one or a group of pursuers to catch one or a group of moving evaders \cite{makkapati2019optimal}. Similarly, the robot soccer domain \cite{nadarajah2013survey} deals with how one group of robots wins over another group of robots on a typical game element.

\begin{figure}[tbp]
        \centering
        \includegraphics[width=0.5\textwidth]{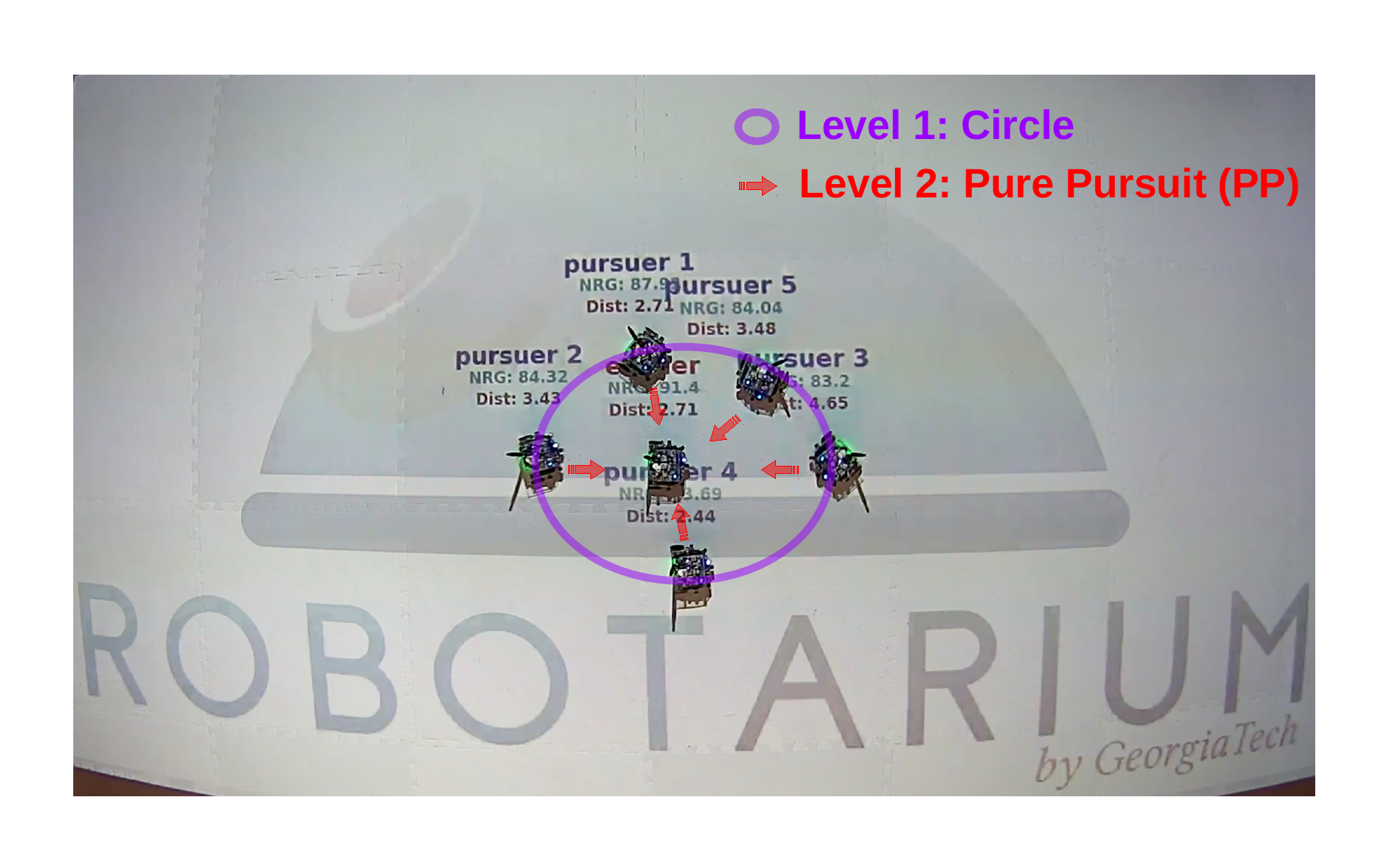}
        % \vspace{-2mm}
        \caption{Illustration of five pursuers catching one evader in Pursuit-Evasion game through the game-theoretic utility tree (GUT) with circle formation (level 1) and PP tactic (level 2) in an experiment using the Robotarium \cite{wilson2020robotarium} platform.}
        \label{fig:gut_explore_domain}
        \vspace{-4mm}
\end{figure}

Furthermore, the {\it Pursuit-Evasion} problems have been studied using a wide variety of approaches and have many different instantiations that can be used to illustrate different MAS scenarios \cite{stone2000multiagent}. Recently, \cite{makkapati2019optimal} presents optimal evader strategies to escape from a pursuer when the pursuers follow either a constant bearing (CB) or a pure pursuit (PP) strategy to capture the evader. Here, each evader is assigned to a set of pursuers based on the instantaneous positions of all the players. More specifically, in the case of a CB strategy, the bearing angle between a pursuer and the evader remains constant until the time of capture. But in the PP strategy, the velocity vector of the pursuer is aligned along the line of sight. 
From the perspective of multi-robot cooperative capture, \cite{kothari2014cooperative} discussed an approximate solution that the pursuers minimize the safe-reachable area (the set of points where an evader can travel without being caught) of the evader. 
However, the literature does not address the problem of how to organize robots' group behaviors, presenting a more complex and efficient strategy to catch the evader. 
Therefore, we make the following contributions.

\textbf{Contributions}
\begin{itemize}
\item Firstly, this paper presents a new hierarchical network-based model, termed Game-theoretic Utility Tree (GUT), to arrive at a cooperative pursuit strategy catching the evaders in the Pursuit-Evasion problem by creating a multi-level strategy decomposition.

\item Secondly, we demonstrate the performance of the GUT in the real robot implementing the GUT algorithm in the Georgia Tech's \textbf{Robotarium} platform \cite{wilson2020robotarium}, which is an open access multi-robot testbed. We open source this implementation in GitHub\footnote{\url{https://github.com/herolab-uga/gut-pursuit-evasion-robotarium}}.

\item Finally, we verify the effectiveness of the GUT through simulations and real-robot experiments by comparing it to the conventional constant bearing (CB) and pure pursuit (PP) strategies. The results show that the GUT could effectively organize cooperation strategies, helping the group with fewer advantages achieve higher performance.

\end{itemize}

Fig. \ref{fig:gut_explore_domain} demonstrates five pursuers catching one evader in the Pursuit-Evasion game through the game-theoretic utility tree (GUT) with circle formation (level 1) and PP tactic (level 2) in Robotarium.

\section{Related Work}

Recently, multiagent systems (MAS) working in adversarial environments like hazardous and disaster scenarios became an emergent topic in robotics, especially implementation in multi-robot systems (MRS). They have four main categories based on task performance: Adversarial Patrol \cite{agmon2009adversarial}; Adversarial Coverage \cite{agmon2011multi,yehoshua2015adversarial}; Adversarial Formation \cite{shapira2015path}; and Adversarial Navigation \cite{keidar2018safe}.
For no artificial agents, they might be a natural force like wind, fire, rain, or obstacles. Many studies solve these problems by detecting opponents, path planning avoiding static or dynamical obstacles, rendezvous and formation control, and so forth \cite{shapira2015path,yang2019self,parasuraman2018multipoint}. 
More specifically, in the urban search and rescue (USAR) missions, artificial agents (like robots) need to face various unintentional adversaries, such as radiation, clutters, or natural forces such as wind and fire, adopting suitable plans and tactics to adapt it \cite{yang2020needs}.

In the game-theoretic applications research, game-theoretic approaches are still a promising new direction to distributed control of MAS. Especially, the optimal control of MAS via game theory assumes a system-level object is given, and the utility functions for individual agents are designed to convert a multiagent system into a potential game \cite{liu2019game}. Specially, \cite{bacsar1998dynamic} studied the dynamic non-cooperative game theory using distributed optimization, \cite{marden2009cooperative} investigated the MAS consensus, and \cite{xu2018nash} provided an algorithm for large-scale MAS optimization. Furthermore, \cite{gopalakrishnan2011architectural} describes multiagent control problems using potential game theory architecture. Moreover, there are still open challenges in the area, such as designing utility functions, learning from global goals for potential game-based optimization of control systems, and converting the local optimization problem to an original optimization problem into a potential networked game \cite{abouheaf2014multi}.

On the other hand, cooperative decision-making among the agents is essential to address the threats posed by intentional physical adversaries or determine tradeoffs in tactical networks \cite{cho2014tradeoffs}. 
Current research mainly focuses on solving multiplayer \textit{pursuit and evasion} game problem \cite{chung2011search, kolling2010multi}, which primarily deals with how to guide one or a group of pursuers to catch one or a group of moving evaders. 
Recent works in this domain concentrate on optimal evasion strategies and task allocation \cite{makkapati2019optimal} and predictive learning from agents' behaviors \cite{shivam2019predictive}.

Furthermore, from the realistic and practical perspective, \cite{yang2020game} design a new game domain called \textit{Explore Domain}, which can analyze how to organize more complex relationships and behaviors in MAS cooperation, achieving given tasks with a higher success probability and lower costs in uncertain environments. In this domain, multiple explorer agents need to cooperatively execute assigned tasks while overcoming the adversarial opponent agents in the environment.

As we can see, organizing complex relationships through a hierarchical structure enabled by GUT will significantly help address the challenges in the multiagent pursuit-evasion problems in the literature.

\section{Background and Preliminaries}
This section provides essential background about \textit{Utility Theory}, {\it Game Theory} and \textit{Game-theoretic Utility Tree (GUT)}. 
We use the notations and the relative definitions from the corresponding papers when describing a specific method.

\subsection{Utility Theory}
The dominant approach to modeling an agent's interests or needs is {\it utility theory}. This theoretical approach aims to quantify an agent's degree of preference across a set of available alternatives and understand how these preferences change when an agent faces uncertainty about which alternative it will receive \cite{shoham2008multiagent}. 

To describe the interactions between multiple utility-theoretic agents, we use the specific {\it utility function} to analyze their preferences and rational action. The utility function is a mapping from states of the world to real numbers, which are interpreted as measures of an agent's level of happiness (needs) \cite{yang2020hierarchical} in the given states. If the agent is uncertain about its current state, the utility is defined as the {\it expected value} (Eq. \eqref{sub_utility1}) of its utility function for the appropriate probability distribution over the states \cite{shoham2008multiagent}. From the perspective of the connection between a decision-maker and its preference, a decision-maker would rather implement a more preferred alternative (act, course of action, strategy) than one that is less preferred \cite{fishburn1970utility}.
\begin{equation}
        \begin{split}
            \mathbb{E} \left[ u(X) \right] = \sum_i u(x_i) \cdot P(x_i)
        \label{sub_utility1}
        \end{split}
\end{equation}

\subsection{Game Theory}
Game Theory is the science of strategy, which provides a theoretical framework to conceive social situations among competing players and produce optimal decision-making of independent and competing actors in a strategic setting \cite{myerson2013game}. In a non-cooperative game, players compete individually and try to raise their profits alone. Especially in the zero-sum games, their total value is constant and will not decrease or increase, which means that one player's profit is associated with another loss. In contrast, if different players form several coalitions trying to take advantage of their coalition, that game will be cooperative \cite{sohrabi2020survey}.

For the game's solutions, if players adopt a \textit{Pure Strategy}, it will provide maximum profit or the best outcome. Therefore, it is regarded as the best strategy for every player of the game. On the other hand, in a \textit{Mixed Strategy}, players execute different strategies with the possible outcome through a probability distribution over several actions.

Furthermore, \textit{Nash Existence Theorem} is a theoretical framework that guarantees the existence of a set of mixed strategies for a finite, non-cooperative game of two or more players in which no player can improve its payoff by unilaterally changing strategy. It guarantees that every game has at least one Nash equilibrium \cite{jiang2009tutorial}, which means that every finite game has a \textit{Pure Strategy Nash Equilibrium} or a \textit{Mixed Strategy Nash Equilibrium}. 

\subsection{Game-theoretic Utility Tree (GUT)}

\begin{figure}[tbp]
\centering
\includegraphics[width=1\columnwidth]{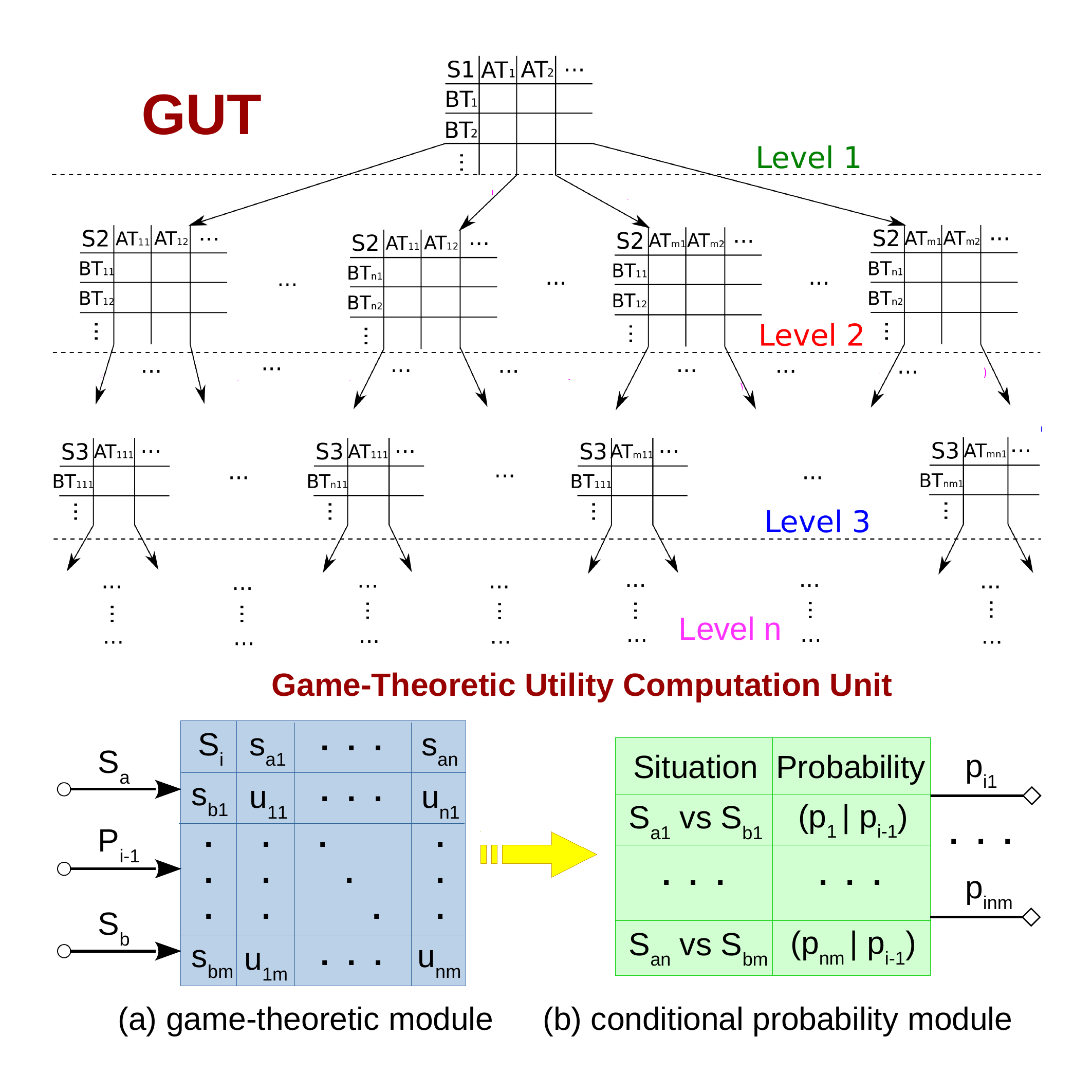}
%\vspace{-2mm}
\caption{Structure of the Game-theoretic Utility Tree (GUT) for hierarchical decomposition of strategies.}
\label{fig:game_decision_trees}
% \vspace{-4mm}
\end{figure}

\textit{Game-theoretic Utility Tree (GUT)} \cite{yang2020game} is a new network model for artificial intelligent agent decision-making or MAS achieving cooperative decisions in uncertain environments, especially in adversarial scenarios. It builds a hierarchical relationship between individual behaviors and interactive entities and helps agents represent more complex behaviors to adapt to various scenarios. Instead of computing an intricate big game involving all the strategies in a complex situation, GUT decomposes the big game (considering all strategy combinations in one game) into several simple dependent games with corresponding strategies representing them as a tree structure to describe the complex situation. It helps reduce the time complexity of finding an optimal strategy combination.

On the other hand, a polynomial-time algorithm for normal-form games is of little use if the normal form is too large for the computer to store, and in this case, the computer needs to operate directly on a more concise representation of the game \cite{conitzer2008new}. By decomposing the big game into simple games and organizing them as a tree, the GUT also improves the space complexity in the entire computation process.

Specifically, \textit{GUT} consists of \textit{Game-theoretic Utility Computation Units} distributed in multiple levels by decomposing strategies, thereby significantly lowering the game-theoretic operations in the strategy space dimension. It combines the core principles of \textit{Utility Theory} and \textit{Game Theory}. Moreover, the payoff (utility) values in the \textit{Game-theoretic Utility Computation Units} are calculated through the agent needs expectations organized hierarchically following the \textit{Agent (robot) Needs Hierarchy} \cite{yang2020hierarchical, yang2021can}. As a comprehensive artificial intelligent architecture, \textit{GUT} not only considers the data from perceiving the environments with different "senses" (e.g., vision and hearing) but also infer the world's conditional (or even causal) relations and corresponding uncertainty with the hierarchical network.

Fig.~\ref{fig:game_decision_trees} outlines the structure of the \textit{Game-theoretic Utility Tree (GUT)} and its computation units distributed in each level. 
First, the \textit{game-theoretic module} (Fig.~\ref{fig:game_decision_trees} (a)) calculates the nash equilibrium based on the utility values $(u_{11}, ... , u_{nm})$ of corresponding situations, $(p_1, ... , p_{nm})$ presenting the probability of each situation.
Then, through the \textit{conditional probability (CP)} module (Fig.~\ref{fig:game_decision_trees} (b)), the CP of each situation can be described as $(p_{i1}, ... , p_{inm})$, where $p_{inm} = (p_{nm} | p_{i-1}),~i,n,m \in Z^+$.

\textbf{Here,} $p_{i-1}$ and $S_i$ present the probability of previous situation and current strategy in the game-theoretic payoff table; $s_a$, $s_b$ and $n$, $m$ represent their strategy space and size on both sides, respectively.

\section{Methodology}
\begin{figure*}[tbp]
        \centering
        \includegraphics[width=1\textwidth]{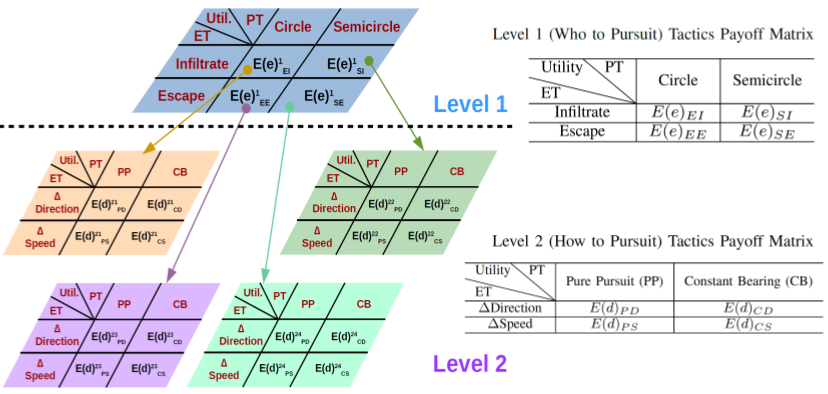}
        % \vspace{-2mm}
        \caption{Illustration of two level game-theoretic utility tree (GUT) for multi-robot cooperative pursuit}
        \label{fig:gut_pursuit}
        % \vspace{-4mm}
\end{figure*}

As we discussed above, agents first decompose the specific group strategy into several independent sub-strategies based on the same category of individual low-level primitives or atomic operations in the GUT \cite{yang2020hierarchical, yang2020game}. Then, through calculating various Nash equilibrium based on different situation utility values in each level's \textit{Game-theoretic Utility Computation Units}, agents can get optimal or suboptimal strategy sets tackling the current status according to \textit{Nash Existence Theorem}. 

From the task execution perspective, the \textit{GUT} can be regarded as a \textit{Task-Oriented Decision Tree}. It decomposes a big game into several sub-games with conditional dependence in each level, which organizes agents' strategies and presents more complex behaviors adapting to adversarial environments. 

We build a two-level GUT decomposing high-level strategies into executable low-level actions for cooperative pursuit strategy decisions in our scenarios. GUT decomposes a complex big game or scenario into conditionally dependent small games or simple situations and presents them as a tree structure. It combines with a new payoff measure based on agent needs \cite{yang2020hierarchical} for real-time strategy games and provides a novel method to organize agents' group behaviors in the {\it Pursuit} domain. By calculating the {\it Nash equilibrium} in various games distributed in corresponding levels, we can theoretically guarantee to find an optimal strategy (tactics) trajectory in the GUT for the current situation based on {\it Nash Existence Theorem}. Fig. \ref{fig:gut_pursuit} illustrates the two-level GUT for multi-robot cooperative pursuit in the {\it Pursuit-Evasion} game.

\begin{algorithm}[t]
\scriptsize
\KwIn{Pursuers' and Evaders' states. ($d$ = average distance between Pursuers and Evaders)}
\KwOut{formation shape $s$; pursuit strategy $t$.}
\caption{Pursuer's Collective Strategy Using \textit{GUT} Model in the \textit{Pursuit-Evasion Game}.}
\label{alg:DP}
%\BlankLine
set state = $"level~one"$; \\
\While{$|d|$ > $\beta$ ($\beta$ is the capture distance)} 
{
    \uIf{state==$"level~one"$}
    {
        Compute the Nash Equilibrium; \\
        Get the most feasible formation shape $s$; \\
        state = $"level~two"$
    }
    \ElseIf{state==$"level~two"$ And s != Null}
    {
        Compute the Nash Equilibrium; \\
        Get the most feasible pursuit strategy $t$; \\
    }
    % \ElseIf{state==$"level~three"$ And s, t != Null}
    % {
    %     Compute the Nash Equilibrium; \\
    %     Get the most feasible number of groups $g$; \\
    % }   
}
% \If{$\beta$ == 0}
% {
%     $s$ = $"Patrol"$; \\
%     $g$ = 1;
% }
\Return $s,t$
\end{algorithm}

In our pursuit-evasion game, we assumed that the energy and distance utility expectation follow the Eq. \eqref{expected_utility}.
\begin{equation}
\begin{split}
E(e_{cir/sem}, \alpha_{\Delta inf/\Delta esc}) = \alpha_{\Delta inf/\Delta esc} d_{cir/sem} \\
E(d_{cb/pp}, \beta_{\Delta dir/\Delta spd}) = \beta_{\Delta dir/\Delta spd} d_{cb/pp}
\label{expected_utility}
\end{split}
\end{equation}
Here, $d_{cir/sem}$ are the distances to the goal point using circle or semicircle methods. $\alpha_{\Delta inf/\Delta esc}$ are the coefficients of evader with the strategies of infiltrating or escaping, respectively.
$d_{cb/pp}$ presents the current distance between pursuer and capture position, executing CB or PP tactic. $\beta_{\Delta dir/\Delta spd}$ describe the coefficients of evader implementing changing direction or speed strategy following normal distributions with different expectations correspondingly. $E_e$ and $E_d$ are the expected energy and distance utilities between the current position and goal point, respectively.

Alg.~\ref{alg:DP} presents the two-level GUT-based strategic decision-making in the "Pursuit Domain." 
Specifically, the first level defines the pursuer's high-level strategies, which are represented as \textit{Circle} and \textit{Semicircle} formation shapes of the pursuer team. Based on the first level decision, they need to decide the specific pursuit strategy in the second level. Here, we assume that pursuers have two basic tactics: constant bearing (CB) and pure pursuit (PP). Correspondingly, the evader also has two categories of strategies -- \textit{Infiltrate} and \textit{Escape} in the first level, \textit{Changing Direction} and \textit{Changing Speed} in the second level -- distributed at different levels, respectively.

\section{Experiments and Evaluation}
\begin{figure*}[t]
%\vspace{-4mm}
\centering
 \subfigure[1 pursuers vs 1 evader]{
    \begin{minipage}[t]{0.314\linewidth}
    \includegraphics[width=1\textwidth]{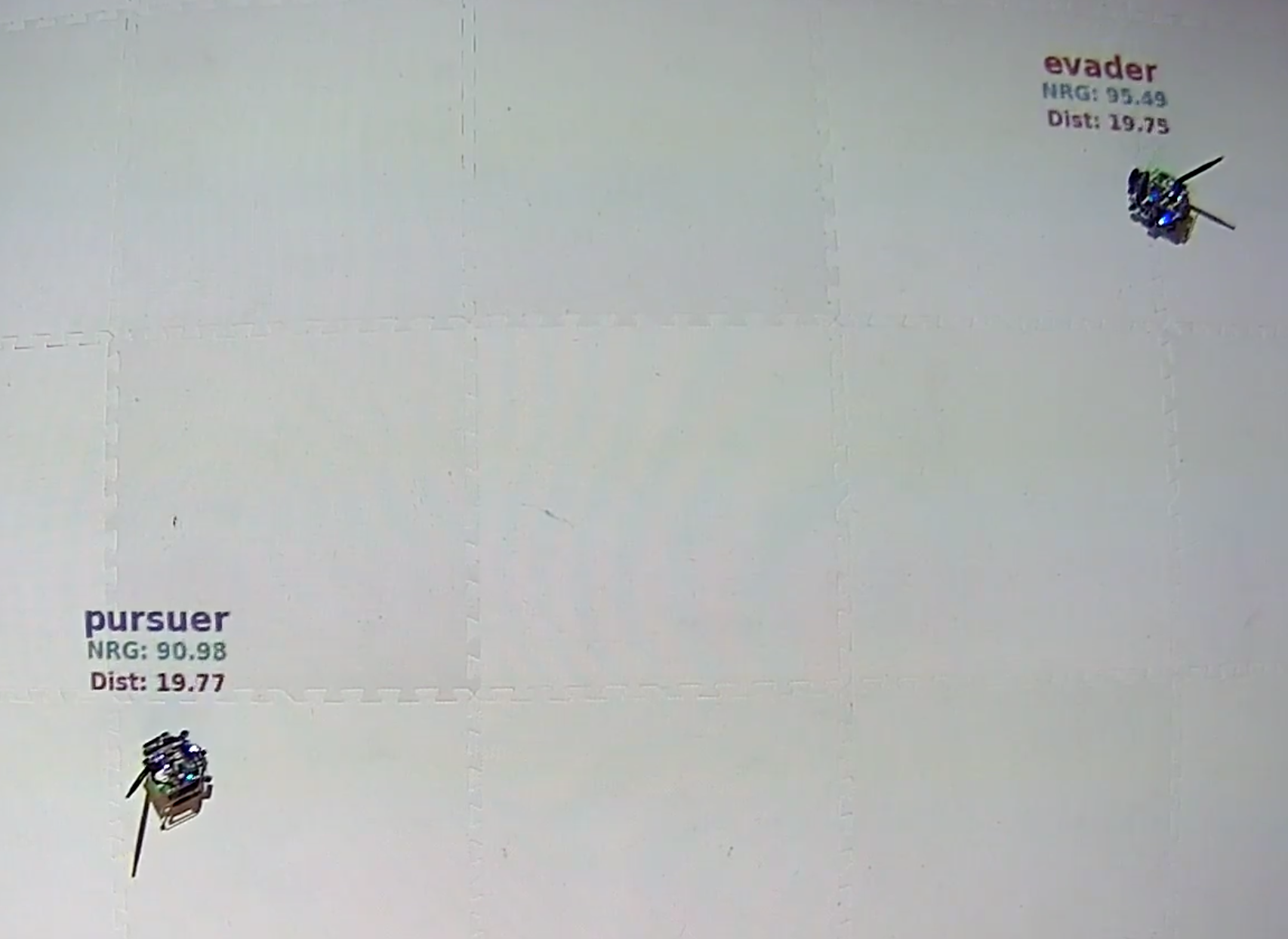}
    \label{fig: 1pvs1e}
    \end{minipage}}
    % \vspace{-2mm}
    \subfigure[Pursuer Average Energy Cost in A Round]{
    \begin{minipage}[t]{0.325\linewidth}
 \includegraphics[width=1\textwidth]{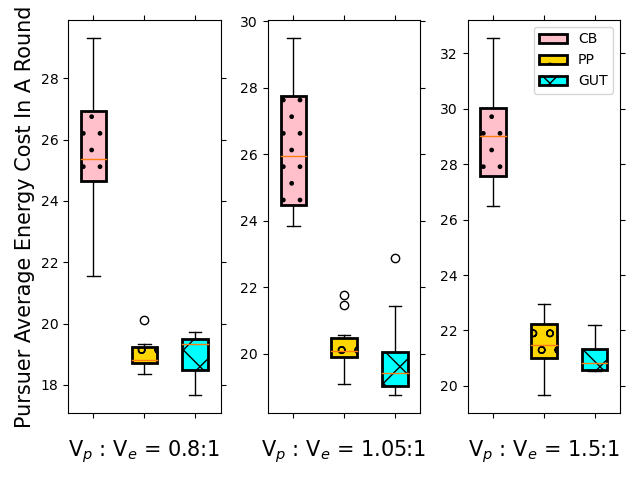}
 \label{fig: pursuit_energy_1v1}
 \end{minipage}}
 \subfigure[Time To Catch Evader Per Round]{
    \begin{minipage}[t]{0.325\linewidth}
 \includegraphics[width=1\textwidth]{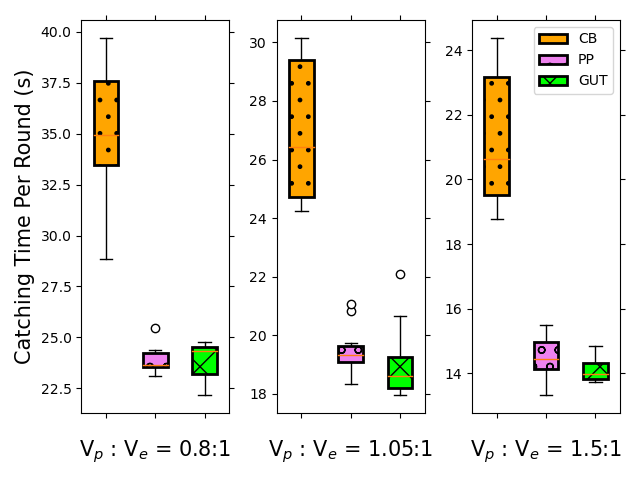}
 \label{fig: pursuit_time_1v1}
 \end{minipage}}
\caption{\small{The Performance (1P vs 1E) of Robotarium Experiments with Different Speed Proportion in Pursuit-Evasion Domain.}}
\label{fig: robotarium_pursuit_experiment_1vs1}
% \vspace{-2mm}
\end{figure*}

\begin{figure*}[h]
%\vspace{-4mm}
\centering
 \subfigure[3 pursuers vs 1 evader]{
    \begin{minipage}[t]{0.314\linewidth}
    \includegraphics[width=1\textwidth]{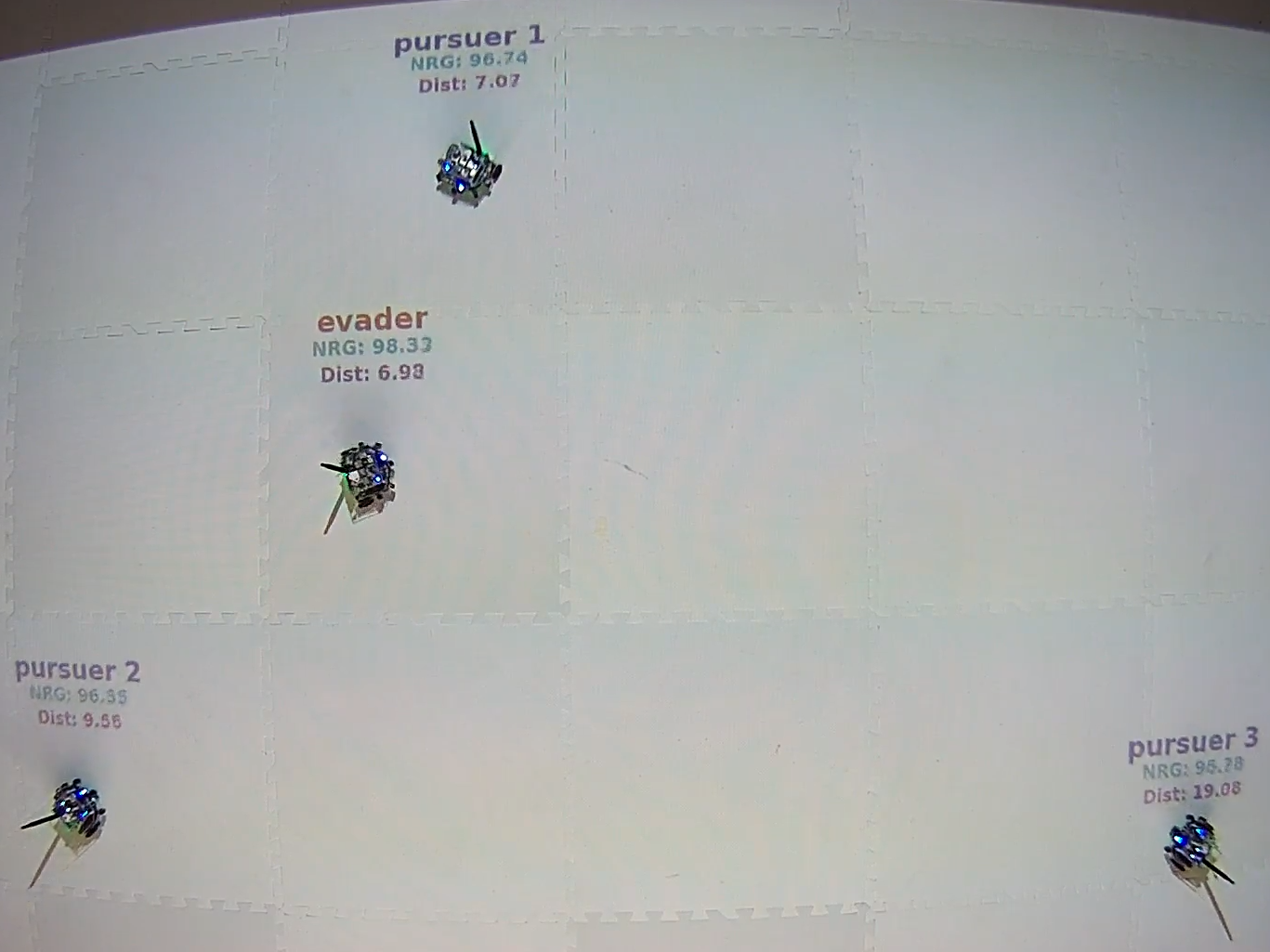}
    \label{fig: 3pvs1e}
    \end{minipage}}
    % \vspace{-2mm}
    \subfigure[Pursuer Average Energy Cost in A Round]{
    \begin{minipage}[t]{0.325\linewidth}
 \includegraphics[width=1\textwidth]{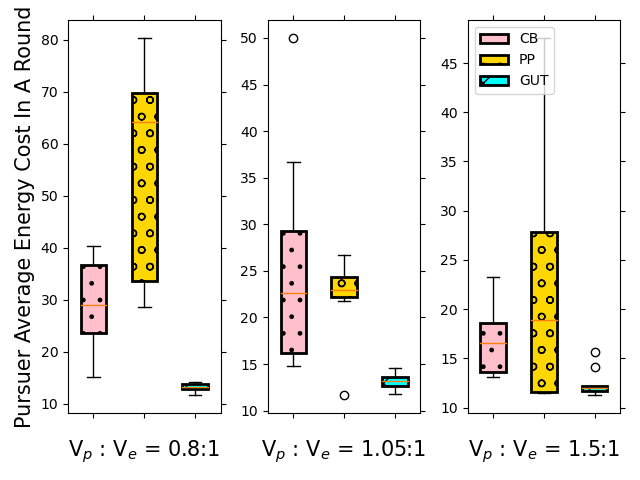}
 \label{fig: pursuit_energy_3v1}
 \end{minipage}}
 \subfigure[Time To Catch Evader Per Round]{
    \begin{minipage}[t]{0.325\linewidth}
 \includegraphics[width=1\textwidth]{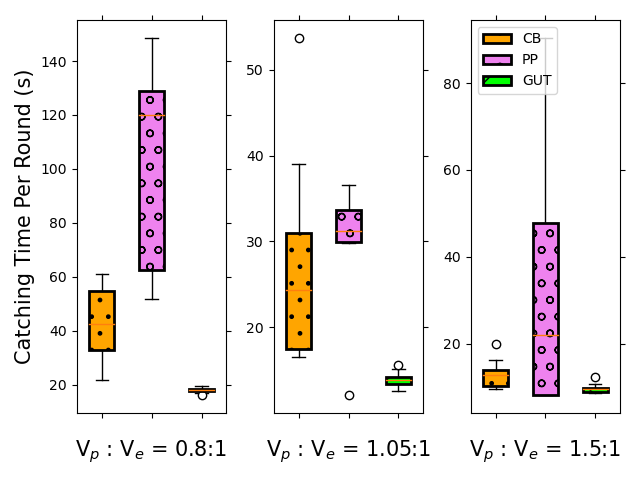}
 \label{fig: pursuit_time_3v1}
 \end{minipage}}
\caption{\small{The Performance (3P vs 1E) of Robotarium Experiments with Different Speed Proportion in Pursuit-Evasion Domain.}}
\label{fig: robotarium_pursuit_experiment_3vs1}
% \vspace{-2mm}
\end{figure*}

\subsection{Experiment Setting}
To demonstrate the GUT on the Pursuit-Evasion domain, we implement our method in the Robotarium \cite{wilson2020robotarium} platform, a remote-accessible multi-robot experiment testbed that supports controlling up to 20 robots simultaneously on a 3.2m $\times$ 2.0m large rectangular area. Each robot has the dimensions 0.11 m $\times$ 0.1 m $\times$ 0.07 m in the testbed. The platform also provides a simulator helping users test their code, which can rapidly prototype their distributed control algorithms and receive feedback about their implementation feasibility before sending them to be executed by the robots on the Robotarium. 

Our experiments design a pursuit-evasion game with several pursuers and one evader implementing different approaches to capture the evader: \textit{constant bearing (CB)}, \textit{pure pursuit (PP)}, and \textit{GUT}.
The uncontrolled evaders use either infiltrate or escape approaches (to avoid getting captured by the pursuers), and they can change their direction or speed randomly (unexpectedly) during the game. The pursuers are controlled by our GUT algorithm, which uses a two-level \textit{Pursuer and Evader Tactics Payoff Matrices} structure. The pursuers calculate the utilities based on the expected energy and distance between the current and goal points. 

Especially, the Level one payoff describes the pursuer and evader, choosing which tactic captures the evader and searches the area correspondingly. Based on the work in \cite{kothari2014cooperative}, we design two kinds of high-level strategies in Level one -- \textit{Circle} and \textit{Semicircle} -- for the cooperative capture. In the \textit{Circle} tactic, pursuers would center on the evader and surround it gradually until the evader can not move in any direction. In the \textit{Semicircle} tactic, the pursuers encompass the evader in a semicircle, forcing it to a dead end. On the other hand, the evader will select the strategies -- \textit{Infiltrate} and \textit{Escape} -- based on the current situation correspondingly in Level one.

According to the Level one results, the Level two payoff will be calculated to execute the low-level implementation of the tactic. Here, we directly implement the methods of \textit{CB} and \textit{PP} \cite{makkapati2019optimal} as the pursuers' tactics. The evader will execute \textit{Changing Direction} ($\Delta$ Direction) or \textit{Changing Speed} ($\Delta$ Speed) adapting to the current scenario. We can further extend these hierarchical levels depending on the application domain and possible low-level decomposition of strategies. Both the levels are presented in Fig.~\ref{fig:gut_pursuit}.

Our experiments assume that the pursuers will capture the evader if all the distances between each pursuer and evader are less than the distance threshold $d_{capture}$. i.e., $\sum_{i} \|P_i - E_j\| \leq d_{capture}, i \in N^j_P$, where $N^j_P$ are the number of pursuers $P_i$ seeking to capture the evader $E_j$. We apply three approaches (CB, PP, and GUT) in the Robotarium platform \cite{wilson2020robotarium} both in simulations and real-robot experiments and analyze the results of ten trials. 
We consider two performance metrics: the pursuer's energy costs and the efficiency (time to capture). 

\subsection{Evaluation}

\begin{figure*}[h]
%\vspace{-4mm}
\centering
 \subfigure[5 pursuers vs 1 evader]{
    \begin{minipage}[t]{0.314\linewidth}
    \includegraphics[width=1\textwidth]{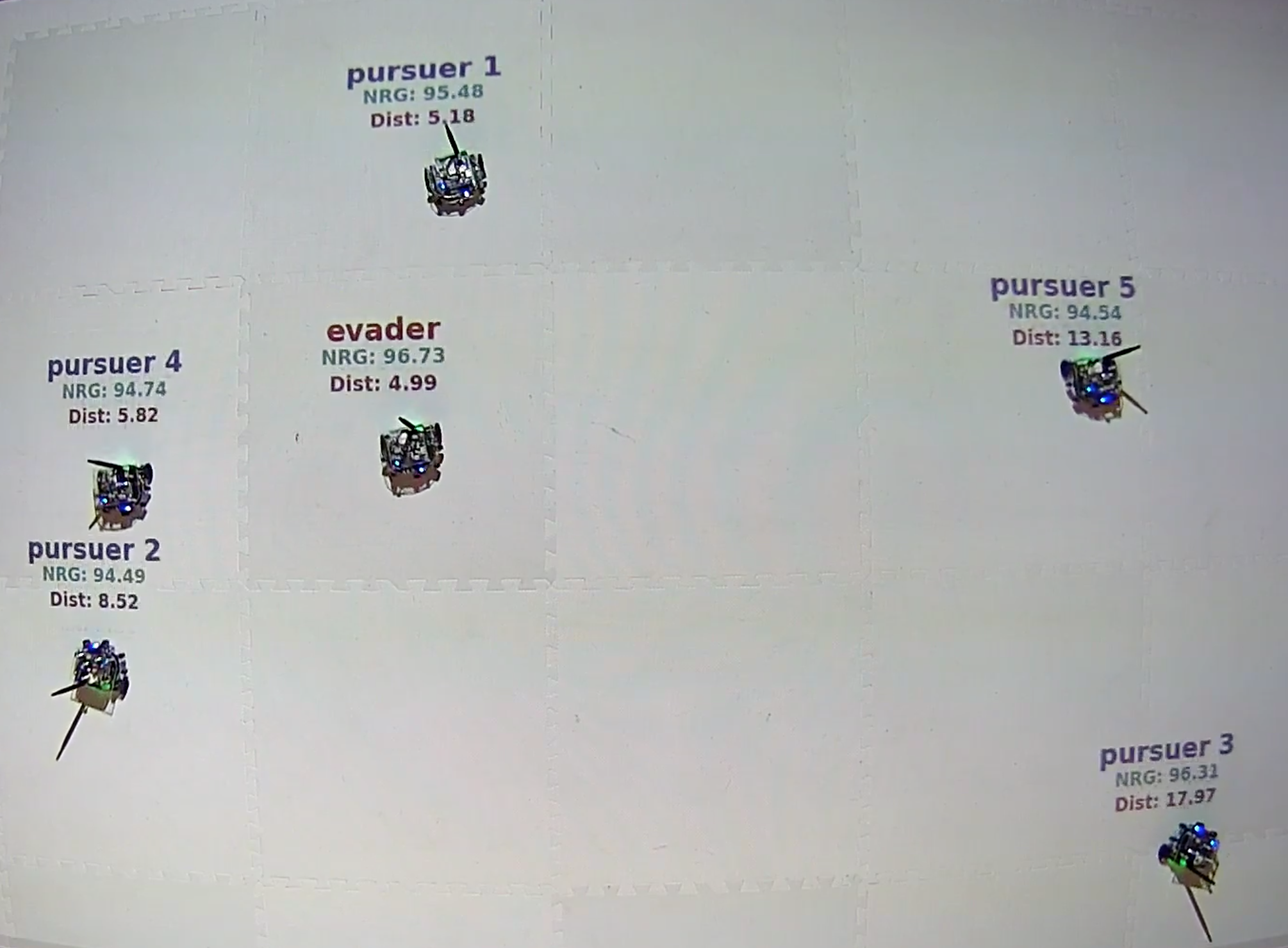}
    \label{fig: 5pvs1e}
    \end{minipage}}
    % \vspace{-2mm}
    \subfigure[Pursuer Average Energy Cost in A Round]{
    \begin{minipage}[t]{0.325\linewidth}
 \includegraphics[width=1\textwidth]{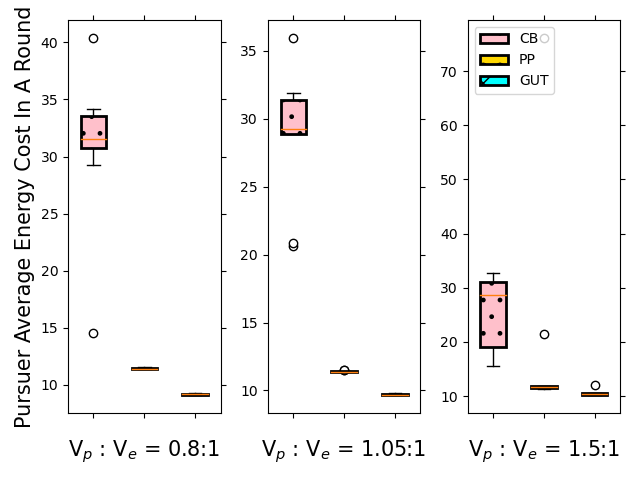}
 \label{fig: pursuit_energy_5v1}
 \end{minipage}}
 \subfigure[Time To Catch Evader Per Round]{
    \begin{minipage}[t]{0.325\linewidth}
 \includegraphics[width=1\textwidth]{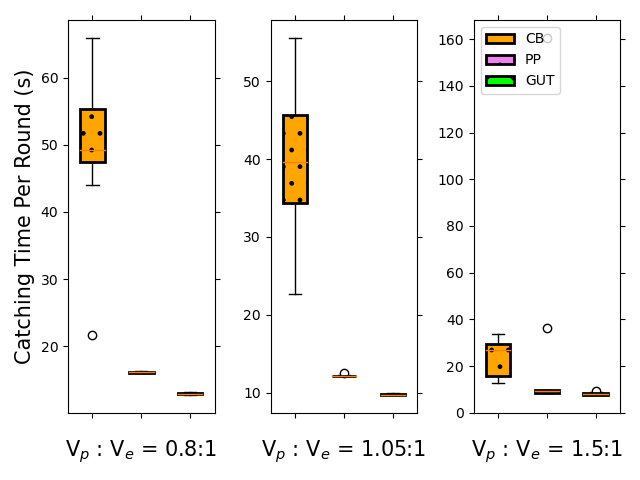}
 \label{fig: pursuit_time_5v1}
 \end{minipage}}
 %\vspace{-4mm}
\caption{\small{The Performance (5P vs 1E) of Robotarium Experiments with Different Speed Proportion in Pursuit-Evasion Domain.}}
\label{fig: robotarium_pursuit_experiment_5vs1}
% \vspace{-2mm}
\end{figure*}

We consider three scenarios with different proportions between pursuer and evader: {\it 1 pursuer vs 1 evader} Fig. \ref{fig: 1pvs1e}, {\it 3 pursuers vs 1 evader} Fig. \ref{fig: 3pvs1e}, and {\it 5 pursuers vs 1 evader} Fig. \ref{fig: 5pvs1e}. In each scenario, we conducted ten simulation trials with different speed proportions ($V_p:V_e$ = 0.8:1, 1.05:1, and 1.5:1) for the GUT against \textit{constant bearing (CB)} and \textit{pure pursuit (PP)} methods by comparing their average energy cost per round and capture time.

\subsection{1 pursuer vs 1 evader}

In this scenario, we consider one pursuer pursuit one evader, Fig. \ref{fig: 1pvs1e}. Fig. \ref{fig: pursuit_energy_1v1} and \ref{fig: pursuit_time_1v1} show that the performance of PP and GUT has a great advantage for CB in every speed proportion. However, comparing the GUT with PP, the pursuer's average energy cost per round and capture time do not show significant difference.

According to our experiments, when we implemented the GUT, the pursuer only selected the PP tactic chasing the evader in this scenario instead of changing its pursuit strategy. It means that GUT will converge to a unique and constant solution in a simple situation without considering the many factors involved in the current utility or reward mechanism. It also implies that the PP strategy might be the optimal solution for this scenario.

\subsection{3 pursuers vs 1 evader}

When we add another two pursuers Fig. \ref{fig: 3pvs1e} in our scenario to capture the evader cooperatively, the results represent distinguished differences among CB, PP, and GUT with different speed proportions.

Fig. \ref{fig: pursuit_energy_3v1} and \ref{fig: pursuit_time_3v1} describe pursuer average energy cost per round and capture time for three different speed proportions, respectively. As we can see, if the pursuer and evader have a similar velocity ($V_p:V_e$ = 1.05:1), the GUT has a more distinct performance than the larger one ($V_p:V_e$ = 1.5:1) compared with CB and PP. Especially when the evader has a more significant advantage than the pursuer ($V_p:V_e$ = 0.8:1), the GUT shows dramatic performance compared with CB and PP.

Furthermore, if there is a considerable advantage in capabilities (like the speed) between individuals and opponents, the performance of cooperation and independence will be similar. Moreover, it is worth the point that suitable strategy selection and change can help the group with fewer advantages achieve higher performance, such as the GUT.
It also verifies that if every group member has outstanding abilities, the cooperation becomes less significant.

\subsection{5 pursuers vs 1 evader}

Considering more pursuers chasing the evader, Fig. \ref{fig: 5pvs1e} presents five pursuers collaboratively capturing one evader in the Robotarium. Fig. \ref{fig: pursuit_energy_5v1} and \ref{fig: pursuit_time_5v1} show that the CB has the worst performance in different speed proportions ($V_p:V_e$ = 0.8:1, 1.05:1, 1.5:1) than others. Furthermore, although the GUT presents some advantages in $V_p:V_e$ = 0.8:1 and 1.05:1 compared with the PP, the performance shows fewer differences in $V_p:V_e$ = 1.5:1. Due to the number of pursuers increasing, the advantages of some strategies become less effective. It means that these strategies which the pursuer can choose have been affected by other factors (like space) in the current dimension.

Through the above \textit{Pursuit} experiments, we demonstrate the effectiveness and generality of the GUT in organizing individual behaviors to present more complex game-theoretic strategies, adapting to various situations to improve system performance. Furthermore, we prove that effective cooperation strategies can help agents with fewer advantages achieve higher performance in the tasks, which effectively organize the system resource and fully take advantage of the capabilities of each group member.
We can further improve the performance of GUT's hierarchically structured relationships by adding more strategies to any levels (dimensions) or combining new strategies.

\section{Conclusions}
Our work extends the Game-theoretic Utility Tree (GUT) in the pursuit domain to achieve multiagent cooperative decision-making in catching an evader. We demonstrate the GUT's performance in the real robot implementing the Robotarium platform compared to the conventional constant bearing (CB) and pure pursuit (PP) strategies. Through simulations and real-robot experiments, the results show that the GUT could effectively organize cooperation strategies, helping the group with fewer advantages achieve higher performance.

In our future work, we plan to improve GUT from different perspectives, such as optimizing GUT structure through learning from different scenarios, designing appropriate utility functions, building suitable predictive models, and estimating reasonable parameters fitting the specific scenario. Besides, optimizing GUT structure through learning from different scenarios with reinforcement learning techniques is also an avenue for future work. Especially, integrating deep reinforcement learning (DRL) into GUT will primarily increase its application areas and effectiveness.

%%% Please leave the following line as is and insert your bibliography items below.

%\def\bibfont{\normalfont\small}
\bibliographystyle{IEEEtran}
\bibliography{references}

\end{document}